\documentclass[conference]{IEEEtran}
\IEEEoverridecommandlockouts

\usepackage{cite}
\usepackage{amsmath,amssymb,amsfonts}
\usepackage{algorithmic}
\usepackage{graphicx}
\usepackage{textcomp}
\usepackage{xcolor}
\usepackage{verbatim} 
\usepackage{multirow}
\usepackage[inline]{enumitem}
\usepackage{tabularx}
\usepackage{balance}
\usepackage{subcaption}
\usepackage{multirow}
\usepackage{booktabs}

\usepackage[labelformat=parens,labelsep=quad,skip=3pt]{caption}

\def\BibTeX{{\rm B\kern-.05em{\sc i\kern-.025em b}\kern-.08em
		T\kern-.1667em\lower.7ex\hbox{E}\kern-.125emX}}
		
	\makeatletter

\def\ps@IEEEtitlepagestyle{%
  \def\@oddfoot{\mycopyrightnotice}%
  \def\@evenfoot{}%
}
\def\mycopyrightnotice{%
  {\footnotesize 978-1-7281-9556-8/21/\$31.00 \textcopyright2021 IEEE \hfill}
  \gdef\mycopyrightnotice{}
}

\begin{document}
	
	\title{Towards On-Device Face Recognition in Body-worn Cameras

	}
	
\author{\IEEEauthorblockN{Ali Almadan and Ajita Rattani}
\IEEEauthorblockA{\textit{Dept. of Elec. Eng. and Computer Science} \\
\textit{Wichita State University, 
Wichita, USA}\\
aaalmadan@shockers.wichita.edu;~ajita.rattani@wichita.edu}

	}
	
	\maketitle
	
	\begin{abstract}
	Face recognition technology related to recognizing identities is widely adopted in intelligence gathering, law enforcement, surveillance, and consumer applications. Recently, this technology has been ported to smartphones and body-worn cameras (BWC). Face recognition technology in body-worn cameras is used for surveillance, situational awareness, and keeping the officer safe. 
	Only a handful of academic studies exist in face recognition using the body-worn camera. A recent study has assembled BWCFace facial image dataset acquired using a body-worn camera and evaluated the ResNet-50 model for face identification. However, for real-time inference in resource constraint body-worn cameras and privacy concerns involving facial images, on-device face recognition is required. To this end, this study evaluates lightweight MobileNet-V2,  EfficientNet-B0, LightCNN-9 and LightCNN-29 models for face identification using body-worn camera. Experiments are performed on a publicly available BWCface dataset. The real-time inference is evaluated on three mobile devices. The comparative analysis is done with heavy-weight VGG-16 and ResNet-50 models along with six hand-crafted features to evaluate the trade-off between the performance and model size. Experimental results suggest the difference in maximum rank-1 accuracy of lightweight LightCNN-29 over best-performing ResNet-50 is \textbf{1.85\%} and the reduction in model parameters is \textbf{23.49M}. Most of the deep models obtained similar performances at rank-5 and rank-10. The inference time of LightCNNs is 2.1x faster than other models on mobile devices. The least performance difference of \textbf{14\%} is noted between LightCNN-29 and Local Phase Quantization (LPQ) descriptor at rank-1.
	In most of the experimental settings, lightweight LightCNN models offered the best trade-off between accuracy and the model size in comparison to most of the models. 
	\end{abstract}

	
\section{Introduction}
Facial recognition technology is used for verifying or identifying individuals from a digital image or a video frame from a video source~\cite{jain2011handbook,ahonen_face_2006}. Face verification determines whether a pair of faces belong to the same identity, while face identification classifies a \emph{probe} face image to a \emph{template gallery} of the specific identity. It is a widely adopted technology for the recognition of identities in surveillance, border control, healthcare, and banking services.

A typical face recognition pipeline consists of face image acquisition, face detection, facial image representation, and matching~\cite{jain2011handbook,ahonen_face_2006}. 
Over the course of time, various facial image representation methods have been proposed for person recognition ranging from holistic such as Principal Component Analysis~(PCA) and Linear Discriminant Analysis~(LDA)~\cite{etemad1997discriminant}, to local image descriptors such as Local Binary Patterns~(LBP)~\cite{ahonen_face_2006} and Histogram of Oriented Gradients (HOG)~\cite{hog_cite}. 
These local image descriptors also called \emph{hand-crafted descriptors or features} have received significant attention over holistic methods due to their lower computational cost and are non-training based (i.e., hand-crafted features can be extracted from a single image without the need of training a feature extraction module).

Deep learning has obtained significant advances in various computer vision applications. Face recognition accuracy has been boosted to an unprecedented level using deep learning~\cite{minaee2019biometric}. Convolutional Neural Networks (CNNs) such as VGG~\cite{vgg_cite}, ResNet~\cite{he2015resnet}, and InceptionNet~\cite{szegedy2015going} have been trained and fine-tuned for face recognition and obtained hallmark accuracy rate of more than $99\%$~\cite{minaee2019biometric}. However, these high-performing networks have enormous space and computational complexity due to the millions of parameters and computations involved~\cite{8763885}. These requirements make the deployment of deep learning models to resource-constrained environments such as mobile phones or wearable devices, challenging. In order to meet the computational requirements of deep learning in resource-constrained environments, a common approach is to leverage cloud computing where deep learning training and inference is performed on the server. However, cloud computing introduces~\emph{high latency and privacy concerns}.

Recently, face recognition technology has been ported to mobile and wearable devices. For instance, Apple introduced the Face-ID moniker in its iPhone X series for mobile user authentication.  Further, recent interest has been drawn on deploying face recognition technology on \emph{body-worn cameras} (BWC) by law enforcement practitioners in order to keep officers safe, enable situational awareness and surveillance~\cite{stanley2015police,blount2015body,bromberg2020public}. 
With $88\%$ of Americans and $95\%$ of Dutch people supporting wearable cameras on police officers, there is strong public support for this technology~\cite{bromberg2020public}. 
Only a handful of work exists on face recognition\footnote{The term face recognition and identification is used interchangeably.} using body-worn cameras~\cite{Al-Obaydy11, bryan2020effects, brown2016enhanced}. Recent study by the authors~\cite{almadan2020bwcface} assembled \emph{BWCFace dataset} consisting of facial images captured using body-worn cameras in in-door and daylight conditions, and evaluated ResNet-50~\cite{he2015resnet} based CNN model for face identification on this data. 

The \emph{real-time inference} is critical for applications involving face recognition using body-worn cameras. Latency and privacy issues are involved in the transfer of facial images to the cloud for network training and inference. Therefore, towards on-device face recognition in body-worn cameras, we evaluate lightweight CNN models in this study in order to evaluate the tradeoff between the model size and the face identification accuracy obtained.

The contributions of this work are as follows:
\begin{itemize}
\item Evaluation of light weight convolutional neural network (CNN) models namely, MobileNet-V2~\cite{sandler2018mobilenetv2}, EfficientNet-B0~\cite{effi_cite}, LightCNN-9 and LightCNN-29~\cite{wu2018light} for face identification via body-worn camera using publicly available BWCFace dataset~\cite{almadan2020bwcface}. 

\item Comparative performance analysis is done with VGG-16~\cite{vgg_cite} and ResNet-50~\cite{he2015resnet} models to understand the trade-off between the model size and the performance obtained. In addition, six hand-crafted features namely, Local Binary Pattern (LBP)~\cite{lbp_cite}, Modified Local Binary Patterns (mLBP)~\cite{mlbp_cite}, Histograms of Oriented Gradients (HOG)~\cite{hog_cite}, Pyramid Histogram of Oriented Gradients (PHOG)~\cite{phog_cite}, Local Phase Quantization (LPQ)~\cite{lpq_cite}, and Local Ternary Patterns (LTP)~\cite{ltp_cite} are also evaluated for performance comparison.

\end{itemize}

This paper is organized as follows: Section II discusses the prior work on face recognition using body-worn cameras. Deep learning-based models evaluated in this study are discussed in section III. Hand-crafted descriptors used for comparative performance evaluation are discussed in section IV. Experimental protocol and the dataset is discussed in section V. Results are discussed in section VI. Conclusions are drawn in section VII.

	\section{Prior Work}

\label{prior_work}
	\par One of the earliest studies on face recognition using body-worn cameras can be dated back to $2011$ by Al-Obaydy and Sellahewa~\cite{Al-Obaydy11}. Authors collected videos from $20$ subjects using an iOPTEC-P300 body-worn digital video camera and have evaluated traditional subspace methods, namely, PCA, LDA, and discrete wavelet transforms, for facial image representation. The cosine similarity metric was used to compute scores between feature vectors from the pair of facial images for the final identity assignment. Rank-1 recognition accuracy values in the range [$65.83\%$, $76.04\%$] was obtained.  
	
 	\par Brown and Fan~\cite{brown2016enhanced} evaluated three face detectors: Viola \& Jones, Aggregate Channel Feature, and Faster R-CNN on an in-house dataset containing $638$ face images captured using the body-worn camera. 
	 R-CNN approach achieved the highest accuracy of $94\%$ over other face detection methods. 
	 The challenges included detecting blurry, dark, and occluded faces captured in an uncontrolled environment.

\par Recently, Almadan et al.~\cite{almadan2020bwcface} assembled a publicly available BWCFace dataset consisting of a total of $178$K facial images from $132$ subjects captured using body-worn cameras in in-door and daylight conditions. BWCFace is the latest publicly available dataset with \emph{largest number of subjects} captured using Body-worn camera till date. ResNet-50~\cite{he2015resnet} model was used for open-set face identification on the collected dataset. Rank-1 to Rank-10 identification accuracy values were used for performance evaluation. The model obtained the highest Rank-1 accuracy of $99\%$ when templates and probes were acquired in daylight. The model obtained the highest Rank-1 accuracy of $95.19\%$ in cross-lighting conditions.

Table~\ref{summ_lit} summarizes the existing literature on facial analysis using the body-worn camera.	
	
	\begin{table}[]
	\caption{Summary of the existing facial analysis studies using the body-worn camera.}
	\begin{center}
	 
\begin{tabular}{lll} \hline
\textbf{Reference} & \textbf{Dataset Size} & \textbf{Method} \\ \hline
Al-Obaydy & 20 subjects  & PCA, LDA    \\
   and Sellahewa~\cite{Al-Obaydy11}                        & x 96 images & and Discrete Wavelet
                                      \\ & (UBHSD dataset) & \\ \hline
Brown and Fan~\cite{brown2016enhanced} & $638$ face images    & Viola \& Jones, \\ 
                                       &                  & Agg. Feature,               \\
                                       &                   &  Faster R-CNN  \\ \hline
Bryan~\cite{bryan2020effects} &    $3,600$ face images           & Neurotechnology SDKs 
\\ \hline
                                       
Almadan et al.~\cite{almadan2020bwcface} & $132$ subjects  & ResNet-50  \\ 
                  & x $178$K                          & for face identification \\
                  & (BWCFace dataset)          \\
                  &  & \\ \hline
\end{tabular}
	\end{center}
	\label{summ_lit}
\end{table}

	\section{CNN Architectures}
	\label{exps}

A Convolutional Neural Network (CNN) is a type of feed-forward artificial neural network in which the connectivity pattern of its neurons,  with shared learnable weights and biases,  is inspired by the organization of the visual cortex  (http://deeplearning.net/tutorial/lenet.html). A CNN consists of an input and an output layer, as well as multiple hidden layers: convolutional layers, pooling layers,  fully connected layers,  and normalization layers. 
We evaluated lightweight; LightCNN, MobileNet, and EfficientNet for face identification using the body-worn camera. ResNet-50 and VGG-16 are used for comparative performance assessment with light CNN models. Deep features are extracted from one of the fully connected layers of these trained CNNs for facial image representation.

Next, we discuss these models as follows:

		\subsection{Light-Weight CNN Models}
\begin{figure}[!t]
		\centering
		\includegraphics[scale=0.45]{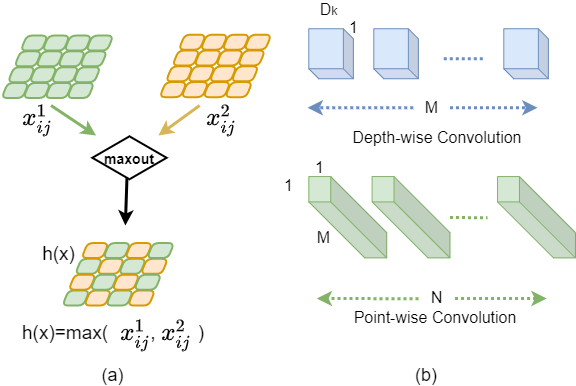}
		\caption{Main component of (a) LightCNN and (b) MobileNet. \cite{wu2018light}}
		\label{fig_cnn}
	\end{figure}

	\begin{itemize} 
	        
	    	\item \textbf{LightCNN}: This model extensively uses the Max-Feature-Map (MFM) operation instead of ReLu activation, which acts as a feature filter after each convolutional layer~\cite{wu2018light}. The operation takes two feature maps ($x^1$ and $x^2$), eliminates the element-wise minimums, and returns element-wise maximums $(h(x)$ as shown in Figure \ref{fig_cnn} (a). By doing so across feature channels, only $50\%$ of the information-bearing nodes from each layer reach the next layer. Consequently, during training, each layer is forced to preserve only compact feature maps. Therefore, model parameters and the extracted features are significantly reduced. We used the lightCNN-$29$ model consisting of \textbf{12.6M} parameters and \textbf{5.5M} parameters for the LightCNN-$9$ model.
	
	        \item \textbf{MobileNet}: MobileNet~\cite{howard2017mobilenets} and~\cite{sandler2018mobilenetv2} is one of the most popular mobile-centric deep learning architectures which is small in size as well as computationally efficient. The main idea of MobileNet is that instead of using regular $3\times3$ convolution filters, the operation is split into depth-wise separable $3\times3$ convolution filters followed by $1\times 1$ convolutions (Figure \ref{fig_cnn} (b)). While achieving the same filtering and combination process as a regular convolution, the new architecture requires a fewer number of operations and parameters. In this study, we used MobileNet-V$2$~\cite{sandler2018mobilenetv2} which consist of \textbf{3.4M} parameters.
	        
	         \item \textbf{EfficientNet}: The network consists of MBConv~\cite{effi_cite} which is similar to the inverted residual blocks in MobileNetv2. This block adds squeeze-and-excitation optimization. The residual blocks are capable of forming a shortcut connection between the beginning of a convolutional block to the end. The total number of parameters of the EfficientNet-B0 is \textbf{5.3M}. 
	        \end{itemize}
	        \subsection{Heavy-weight CNNs Models Used for Comparison}
	        \begin{itemize}
	        
         \item \textbf{ResNet-50}: ResNet~\cite{he2015resnet} is a short form of residual network based on the idea of “identity shortcut connection” where input features may skip certain layers. The residual or shortcut connections introduced in ResNet allow for identity mappings to propagate around multiple nonlinear layers, preconditioning the optimization and alleviating the vanishing gradient problem. In this study, we used ResNet-50 model which has \textbf{23.5M} parameters.
	    \item \textbf{VGG-16}: VGG-16~\cite{vgg_cite}  architecture   consists   of   sequentially   stacked $3 \times 3$ convolutional  layers  with  intermediate  max-pooling  layers followed  by  a  fully  connected  layers  for  feature extraction. The VGG-16 model has a total of \textbf{138M} parameters.   
	   
	    \end{itemize}

	\section{Handcrafted Descriptors}
	These hand-crafted features reported in the literature have been manually designed, or “handcrafted” for feature representation from an image using pixel intensity comparison or gradient information.
		The handcrafted features used in this study are discussed below. 

	\begin{itemize}
    	    \item \textbf{LBP}: Local Binary Pattern (LBP)~\cite{lbp_cite} is a low computational, efficient descriptor that thresholds the neighbors of each pixel in an image. LBP identifies local regions by
    	    considering a N x N size window to calculate the subtraction between the central pixels and their neighboring pixels. If the resulting intensity is greater or equal to one, then the local region is set to 1; otherwise, it is set to zero. The binary code is computed for each region and converted into a decimal number and the histogram of the frequencies is computed and used as a feature vector.   
    	    
    	    \item \textbf{mLBP}: Modified Local Binary Patterns (mLBP)~\cite{mlbp_cite} compares the neighboring pixels to the average of values in the N x N window, instead of the value of the center pixel by LBP.
    	    
    	    \item \textbf{HOG}: Histograms of Oriented Gradients (HOG)~\cite{hog_cite} can re-represent the shape of a local object by using its distribution of edge directions without regards to the spatial distribution of the image. Its advantage over other descriptors is that it is variant only to object orientation. 
    	    
    	    \item \textbf{PHOG}: Pyramid Histogram of Oriented Gradients (PHOG)~\cite{phog_cite} is an extension to the HOG descriptors. To preserve the spatial layout of the image, the PHOG divides a given image into different cells at various pyramid levels. In contrast, HOG operates at a fixed size image. 
    	    
    	    \item \textbf{LPQ}: Local Phase Quantization (LPQ)~\cite{lpq_cite} is a LBP variant that is insensitive to image blur. LPQ utilizes the blur invariance attribute of the Fourier phase spectrum to extract the local phase of an image.  
    
    	    \item \textbf{LTP}: Local Ternary Patterns (LTP)~\cite{ltp_cite} is another variant of LBP which is powerful against noise; however, LTP may be sensitive to illumination change. The LTP computes three-level thresholding, unlike LBP with two-level thresholding. 
	\end{itemize}
These handcrafted features are used for facial feature extraction. We used block size of $32$ and $3\times 3$ window size for handcrafted feature extraction. These features are used for comparative performance analysis with lightweight CNN models.

\section{Experimental Dataset and Protocol}
In this section, we describe BWCFace dataset and the experimental protocol used in this study.

\textbf{BWCFace Dataset}~\cite{almadan2020bwcface}: This dataset is collected from students at Wichita State University using the Body-worn Camera (BWC)~\cite{almadan2020bwcface}. For the data collection, videos from $132$ subjects were collected from Boblov $1296$P WiFi Body Mounted Camera\footnote{https://www.boblov.com/}. The subjects were recorded using a $30$s high definition video of resolution $1920\times1080$ pixels at $30$fps in in-door and daylight settings ($10$ to $15$ minutes apart). The body-worn camera was mounted on the center chest of the Research Assistant following the recommendation in~\cite{bryan2020effects}. During recording, subjects were asked to act naturally in order to represent realistic conditions. The recording was done in an uncontrolled environment with non-uniform background and varying distances between the subject and the body-worn camera. 

Ethnic groups of the captured subjects include Asian, White American, Black, and Middle Eastern. After the data collection, frames were extracted from the collected videos, and facial regions were detected and cropped using Dlib library~\cite{king2009dlib}.
\par Most of the images in BWCface are blurry or have a very low resolution because of the shooting environment and the specification of the camera device, such as not having autofocus. Also, the camera records at a wide angle of 170 $^{\circ}$ for clear and wide footage to cover the whole scene for evidence, unlike mugshots or selfie images for the task of face recognition.
In fact, the camera holder and the target subjects are not stable while shooting in an un-collaborative environment. 

\par The selection of images was random from a large set of images at each specific lighting condition. For instance, the chosen subsets of "Day vs. Office" may not be overlapping with the subset for "Office vs. Day". Therefore, there might be a drop in performance for different scenarios. 

\par The CNN models used in this study are trained from scratch on MS1Mv2~\cite{deng_arcface:_2019} dataset.  
	The MS1Mv2 dataset~\cite{deng_arcface:_2019} is a cleaned version of the MS1M dataset~\cite{guo2016ms}, containing around $5.8$ million images from $85,742$ subjects were used for training the models. 
	For the purpose of experiments, all the images were resized to $224\times 224$. 
	The models were trained using an early stopping mechanism using Adam optimizer and cross-entropy loss function.

For the open-set evaluation of the trained deep learning models for face identification on BWCFace dataset, the template gallery set consists of $12$ face images per subject. The probe set consists of $100$ face images randomly selected per subject.  From the gallery and probe set of all the subjects, deep features of size $512$ are extracted for each image from the fully connected layer of the CNN architectures discussed in Section III.
The matching score between the deep features from the pair of gallery and probe image is computed using cosine similarity given in eq~\ref{cos_sim}:

\begin{equation}
\label{cos_sim}
    	\operatorname{sim}(u, v)
= \frac{u \cdot v}{|u||v|}
= \frac{\sum_{i = 1}^N u_i v_i}
{\sqrt{\left(\sum_{i = 1}^N u_i^2\right)
		\left(\sum_{i = 1}^N v_i^2\right)}}
\end{equation}
	where $u$ and $v$ are the two deep feature vectors: $u = \{u_1,u_2,\dots,u_N\}$ 
	and $v = \{v_1,v_2,\dots,v_N\}$.

The evaluation of handcrafted features follows the same protocol. The matching scores of handcrafted features from a pair of gallery-probe face images were calculated using Euclidean distance given in eq~\ref{euc_dist}.
	
\begin{equation}
\label{euc_dist}
     d\left( u,v\right)   = \sqrt {\sum _{i=1}^{N}  \left( v_{i}-u_{i}\right)^2 }
\end{equation}
	where $u$ and $v$ are the handcrafted feature vectors.

The PyTorch framework was used for CNN training and feature extraction and MATLAB was used for handcrafted feature extraction. The average of the scores from the multiple gallery images per probe is used for the final identity assignment. In cross-lighting conditions, the template (gallery set) was selected from a different lighting condition with that of probe set for all the identities.
\begin{table}[]
 \caption{Real-time Inference: Deep feature extraction time per sample measured in milliseconds (ms) on three real mobile devices for each CNN model.}
\label{my-label}
\begin{center}
   \begin{tabular}{lllc}
    \toprule
    \multicolumn{1}{c}{{CNN Model}} & \multicolumn{1}{c}{{Num. of Params (M)}} & \multicolumn{1}{c}{{Device}} &  \multicolumn{1}{c}{Extraction Time (ms)}  \\

    \midrule
    ResNet-50& \qquad 23.5 & iPhone 6 & \ 2386 \\
                         &  &iPhone X & \ 941 \\
                         &  &iPhone XR & \ 834 \\

     \midrule
    VGG-16& \qquad 138 & iPhone 6 & memory constraints \\
                         & & iPhone X & \ 4138 \\
                         & & iPhone XR & \ 3747 \\

          \midrule
    MobileNetV2& \qquad 3.4 & iPhone 6 & \ 1022 \\
                         &  &iPhone X & \ 357 \\
                         &  &iPhone XR & \ 240 \\

       \midrule                  
    EfficientNet-B0& \qquad 5.3 &  iPhone 6 & \ 1321 \\
                     &  &iPhone X & \ 450 \\
                     &  &iPhone XR & \ 329 \\

       \midrule                  
    LightCNN-29& \qquad 12.6 & iPhone 6 & \ 1426 \\
                     &  &iPhone X & \ 604 \\
                     &  &iPhone XR & \ 430 \\

       \midrule                  
    LightCNN-9& \qquad 5.5 & iPhone 6 & \ 562 \\
                     & & iPhone X & \ 248 \\
                     & & iPhone XR & \ 171 \\

    \bottomrule
    \label{tab:inf_time}
  
    \end{tabular}
      \end{center}
    
\end{table}

\begin{table}[]
\caption{Face identification accuracy of deep CNN models on BWCFace dataset~\cite{almadan2020bwcface} in same and cross-lighting conditions at rank-1, rank-5 and rank-10.}
\begin{center}
\begin{tabular}{|c|c|c|c|}

\hline
\textbf{Light Condition} & \multicolumn{1}{c|}{\textbf{Rank-1 [\%]}} & \multicolumn{1}{c|}{\textbf{Rank-5 [\%]}} & \textbf{Rank-10 [\%]} \\ \hline
	\multicolumn{4}{|c|}{\textbf{LightCNN-9}}                                                                                      \\ \hline
	Office vs. Office            &  93.35                               &  97.78                           & 98.56           \\ \cline{1-1}
	Day. vs. Day                 &    78.70                             &  90.47                             &   93.76           \\ \cline{1-1}
	Day vs. Office               &  22.24                               &   35.63                           & 50.65        \\ \cline{1-1}
	Office vs. Day               & 22.41                                &    32.36                             & 44.75           \\ \hline
	\multicolumn{4}{|c|}{\textbf{LightCNN-29}}                                                                                  \\ \hline
 Office vs. Office            & \textbf{94.48}                              & \textbf{98.27}                                &  \textbf{99.17}           \\ \cline{1-1}
	Day. vs. Day                 & 78.07                                 &   89.41                            &  93.52            \\ \cline{1-1}
	 Day vs. Office               &  22.82                               &  32.86                             & 46.29             \\ \cline{1-1}
Office vs. Day               &      22.93                              &   37.32                               & 47.18            \\ \hline
	\multicolumn{4}{|c|}{\textbf{MobileNet-V2}}                                                                                       \\ \hline
Office vs. Office            &   85.50                              &     97.38                           & 98.99           \\ \cline{1-1}
	Day. vs. Day                 &   72.43                              &    93.98                             &  97.72           \\ \cline{1-1}
	Day vs. Office               &  25.18                               &   33.30                            &  44.60          \\ \cline{1-1}
	Office vs. Day               &   22.35                          &   32.30                             &  44.60           \\ \hline
	
		\multicolumn{4}{|c|}{\textbf{EfficientNet-B0}}                                                                                       \\ \hline
Office vs. Office            &   80.95                          &   94.93                         & 97.61   \\ \cline{1-1}
	Day. vs. Day                 &  68.24                              &   91.18                         &  96.13          \\ \cline{1-1}
	Day vs. Office               & 27.23                            &   42.23                          &   56.08        \\ \cline{1-1}
	Office vs. Day               &    27.92                     &     41.92                  &   55.59        \\ \hline
	\multicolumn{4}{|c|}{\textbf{VGG-16}}                                                                                       \\ \hline
Office vs. Office            &      81.65                           &  95.32                           &  97.71   \\ \cline{1-1}
	Day. vs. Day                 &       69.49                          & 88.73                           &   94.53          \\ \cline{1-1}
	Day vs. Office               &      31.43                          &  49.92                           &   68.08         \\ \cline{1-1}
	Office vs. Day               &     27.22                       &       44.48                         &  58.34           \\ \hline
		\multicolumn{4}{|c|}{\textbf{ResNet-50~\cite{almadan2020bwcface}}}                                                                                        \\ \hline
Office vs. Office            &   {96.33}                              &   {98.47}                          & {99.00}    \\ \cline{1-1}
	Day. vs. Day                 &   \textbf{99.36}                              &  \textbf{99.81}                          &  \textbf{99.85}           \\ \cline{1-1}
	Day vs. Office               &  84.25                              &   95.61                          &    98.06        \\ \cline{1-1}
	Office vs. Day               &    94.45                        &    98.99                            &   99.63          \\ \hline
\end{tabular}
	\end{center}
\label{deep_table}
\end{table}

\begin{table}
\caption{Face identification accuracy of the handcrafted descriptors on BWCFace dataset~\cite{almadan2020bwcface} for same and cross-lighting conditions at rank-1, rank-5 and rank-10.}
\centering
\begin{tabular}{|c|c|c|c|} 
\hline
\textbf{Light Condition} & \textbf{Rank-1 [\%]} & \textbf{Rank-5 [\%]} & \textbf{Rank-10 [\%]}  \\ 
\hline
\multicolumn{4}{|c|}{ \textbf{LBP}}                                                             \\ 
\hline
Office vs. Office        & 60.05                & 83.20                & 90.61                  \\
Day. vs. Day             & 55.13                & 82.18                & 91.01                  \\
Office vs. Day           & 16.66                & 44.07                & 62.16                  \\
Day vs. Office           & 14.55                & 40.22                & 56.65                  \\ 
\hline
\multicolumn{4}{|c|}{ \textbf{mLBP}}                                                            \\ 
\hline
Office vs. Office        & 56.33                & 81.05                & 88.53                  \\
Day. vs. Day             & 51.82                & 80.37                & 89.87                  \\
Office vs. Day           & 16.79                & 43.87                & 62.57                  \\
Day vs. Office           & 16.18                & 41.86                & 56.75                  \\ 
\hline
\multicolumn{4}{|c|}{ \textbf{HOG}}                                                             \\ 
\hline
Office vs. Office        & 65.90                & 84.26                & 89.51                  \\
Day. vs. Day             & 70.39                & 89.46                & 94.12                  \\
Office vs. Day           & 20.64                & 47.31                & 64.01                  \\
Day vs. Office           & 17.96                & 45.67                & 63.04                  \\ 
\hline
\multicolumn{4}{|c|}{ \textbf{PHOG}}                                                            \\ 
\hline
Office vs. Office        & 55.10                & 76.71                & 84.31                  \\
Day. vs. Day             & 64.09                & 84.44                & 91.20                  \\
Office vs. Day           & 11.38                & 33.10                & 45.47                  \\
Day vs. Office           & 10.49                & 32.86                & 49.73                  \\ 
\hline
\multicolumn{4}{|c|}{ \textbf{LPQ}}                                                             \\ 
\hline
Office vs. Office        & \textbf{80.06}                & \textbf{92.29}                & \textbf{95.48}                  \\
Day. vs. Day             & 77.77                & 91.78                & 95.68                  \\
Office vs. Day           & 28.67                & 56.71                & 69.71                  \\
Day vs. Office           & 26.59                & 54.82                & 69.92                  \\ 
\hline
\multicolumn{4}{|c|}{ \textbf{LTP}}                                                             \\ 
\hline
Office vs. Office        & 75.41                & 90.26                & 93.99                  \\
Day. vs. Day             & 10.44                & 32.06                & 45.69                  \\
Office vs. Day           & 10.85                & 33.31                & 47.49                  \\
Day vs. Office           & 10.92                & 38.29                & 51.94                  \\
\hline
\end{tabular}
\label{hcf_table}
\end{table}

\section{Experimental Results}
  Table~\ref{deep_table} shows the face identification accuracy using deep features extracted from the CNN architectures detailed in section III. The deep features have been evaluated in the same and across lighting conditions. For instance, office vs. day represents cross lighting conditions in which templates are acquired in the office light and probe images are acquired in daylight condition.
  
  At office vs. office lighting conditions,  
  both LightCNNs scored an average of 93.92\% at rank-1, followed by $85.50\%$ for MobileNet-V2 at the same rank. VGG16 and EfficientNet-B0 obtained rank-1 accuracy of 81.65\% and 80.95\%, respectively. 
  The highest rank-1 accuracy of $96.33\%$ was obtained by ResNet-50 over all the models.  However, all the models performed similarly at rank-5 and rank-10 ranging from $97.38\%$ to $99.17\%$ accuracy.

    For the day vs. day, the identification accuracy of the ResNet-50 model at rank-1 was higher than VGG16 and EfficientNet-B0 by about $14.68\%$, MobileNet-V2 by $10.83\%$ points, and than both LightCNNs by $2.42\%$ points. However, ResNet-50 and LightCNN-29 had almost similar results at rank-5 and rank-10 with a minimal difference in the accuracy of $0.2\%$. 
    
   The difference in rank-1 accuracy of best-performing lightweight LightCNN over best performing ResNet-50 is $4.88\%$ and the drop in model parameters is $23.49M$. The overall difference in the performance of LightCNN-9 over LightCNN-29 is $0.39\%$ at rank-1 and rank-10, and $0.41\%$ at rank-5. The difference in the model parameters is $7.1K$.
  
   However, for all the deep models, performance degraded significantly across lighting conditions. Across lighting, a maximum accuracy of $94.45\%$ at rank-1 was obtained by ResNet-50 in the day vs. office conditions followed by EfficientNet-B0 and VGG-16 obtaining $27.57\%$ at the same rank and lighting condition. 

   ResNet-50 model obtained a least performance drop between same-lighting and cross-lighting of $8.5\%$, $1.84\%$, $0.58\%$ at rank-1, rank-5, and rank-10, respectively. 
   The accuracy dropped by about $5\%$ for lighter models in cross-lighting condition.
 
    Table \ref{hcf_table} shows the face identification accuracy obtained by handcrafted descriptors (LPB, mLPB, HOG, PHOG, LPQ, and LTP) for same-lighting and cross-lighting conditions. 
     It can be observed that the best rank-1 identification accuracy of $80.06\%$ is obtained using the LPQ descriptor, while the accuracy of $75.41\%$ was obtained using the LTP descriptors. 
     The rest of the descriptors scored similarly across all ranks with an average of $59.34\%$, $81.31\%$, and $88.24\%$ at rank-1, rank-5, and rank-10, respectively. LTP and LPQ descriptors scored the highest with a mean accuracy of $91.28\%$ at rank-5 and $94.73\%$ at rank-10.
    For hand-crafted features, the average decrease in accuracy in cross-lighting conditions is approximately $43.40\%$, $37.99\%$, $29.21\%$ at rank-1, rank-5, and rank-10, respectively. 

    \par In summary, the difference in maximum rank-1 accuracy of highest performing lightweight LightCNN-29 over ResNet-50 and VGG-16 is $-1.85\%$ and $+12.83\%$. The reduction in the model parameters of lightCNN-29 over ResNet-50 and VGG-16 are $23.49M$ and $137.98M$, respectively. The difference further reduces to $1.54\%$ and $8.5\%$, and to $0.68\%$ and $1.46\%$, at rank-5 and rank-10, respectively. 
     Further, the difference in maximum rank-1 accuracy of lightweight LightCNN-29 over best performing LPQ is $14.42\%$ and $5.74\%$ in the same and cross-lighting condition. 
     This difference reduces to $5.98\%$ and $3.69\%$ at rank-5 and rank-10, respectively. The \emph{superior performance of LPQ could be due to its blur invariance property and the face images captured using the body-worn camera contain motion blur}.
     The performance of most of the models and feature descriptors dropped significantly in cross-lighting conditions. ResNet-50 obtained the best overall performance.
     Among, LightCNN-29 and LightCNN-9, the difference in the model parameters is $7.1K$ and the accuracy difference is $0.39\%$ at rank-1. The \emph{superior performance of LightCNN over other lightweight models could be attributed to the use of Max-Feature-Map operation, instead of ReLu, which results in compact and optimized deep feature representation}.

    \par Table \ref{tab:inf_time} shows the real-time inference in terms of feature extraction time per sample on three iPhone devices and the testing environment is Xcode version 12.4. LightCNN-9 obtained the least feature extraction time being 2.1x faster across the three iPhones than EfficientNets. EfficientNets obtained the second lowest execution time although being 1.5x larger than MobileNets. This is due to the efficient, systematic scaling up of the depth, width, and regulation of CNN models.
    \par These results suggest LightCNN models offer the best trade-off between model size and the recognition performance for on-device face recognition in body-worn cameras.

\section{Conclusion}
This paper evaluates different convolutional neural networks (CNNs) and six handcrafted descriptors for face identification using a body-worn camera using a publicly available BWCFace dataset. Deep features are extracted using light and heavy CNN models to measure the trade-off between the model size and identification performance obtained in order to facilitate on-device face recognition. Real-time inference is evaluated on three mobile devices. The experimental results suggest the efficacy of LightCNN models in offering the best tradeoff between the model size (being 2.1x faster) and the recognition performance for the body-worn camera. This could be attributed to the use of Max-Feature-Map operation in LightCNN models which acts as a feature filter in obtaining compact and optimized deep feature representation. Mostly, a large performance difference is noted between lightweight CNNs and the handcrafted descriptors, especially at rank-1. LPQ obtained superior performance over other descriptors which could be due to its robustness to motion blur present in facial images captured using the body-worn camera. As a part of future work, custom compact size CNN models will be developed and evaluated against LightCNN models and ResNet-50, for on-device face recognition using the body-worn camera.

	\section*{Acknowledgment}
	This work is supported in part by Award for Research/Creative projects by Wichita State University. BWCFace dataset is available by contacting the authors.

\end{document}